\documentclass[sn-mathphys]{sn-jnl}

\usepackage{amssymb}
\usepackage{amsthm}
\usepackage{bm}

\usepackage{amsmath}
\usepackage{subfigure}
\usepackage{algorithm}
\usepackage{algpseudocode}

\usepackage{url}

\jyear{2023}%

\theoremstyle{thmstyleone}%
\theoremstyle{thmstyletwo}%

\theoremstyle{thmstylethree}%

\begin{document}

\title[IDRL]{Classifying Ambiguous Identities in Hidden-Role Stochastic Games with Multi-Agent Reinforcement Learning}


\author[1]{\fnm{Shijie} \sur{Han}}\email{22S003048@stu.hit.edu.cn}

\author*[1]{\fnm{Siyuan} \sur{Li}}\email{siyuanli@hit.edu.cn}

\author[2]{\fnm{Bo} \sur{An}}\email{boan@ntu.edu.sg}

\author[1]{\fnm{Wei} \sur{Zhao}}\email{zhaowei@hit.edu.cn}

\author[1]{\fnm{Peng} \sur{Liu}}\email{pengliu@hit.edu.cn}

\affil*[1]{\orgdiv{School of Computer Science and Technology}, \orgname{Harbin Institute of Technolog}, \orgaddress{\street{ 92 West Dazhi Street}, \city{Harbin}, \postcode{150001}, \country{China}}}

\affil[2]{\orgdiv{School of Computer Science and Engineering}, \orgname{Nanyang Technological University},  \country{Singapore}}



\abstract{Multi-agent reinforcement learning (MARL) is a prevalent learning paradigm for solving stochastic games. In most MARL studies, agents in a game are defined as teammates or enemies beforehand, and the relationships among the agents (i.e., their \textit{identities}) remain fixed throughout the game.
However, in real-world problems, the agent relationships are commonly unknown in advance or dynamically changing.
Many multi-party interactions start off by asking: who is on my team? This question arises whether it is the first day at the stock exchange or the kindergarten. 
Therefore, training policies for such situations in the face of imperfect information and ambiguous \textit{identities} is an important problem that needs to be addressed. In this work, we develop a novel identity detection reinforcement learning (IDRL) framework that allows an agent to dynamically infer the identities of nearby agents and select an appropriate policy to accomplish the task. In the IDRL framework, a relation network is constructed to deduce the identities of other agents by observing the behaviors of the agents. A danger network is optimized to estimate the risk of false-positive identifications. Beyond that, we propose an intrinsic reward that balances the need to maximize external rewards and accurate identification. After identifying the cooperation-competition pattern among the agents, IDRL applies one of the off-the-shelf MARL methods to learn the policy. To evaluate the proposed method, we conduct experiments on \textit{Red-10} card-shedding game, and the results show that IDRL achieves superior performance over other state-of-the-art MARL methods. Impressively, the relation network has the par performance to identify the identities of agents with top human players; the danger network reasonably avoids the risk of imperfect identification. The code to reproduce all the reported results is available online at
\url{https://github.com/MR-BENjie/IDRL}.}

\keywords{Multi-agent reinforcement learning, Game theory, Imperfect information, Cooperation-competition, Ambiguous identities}



\maketitle

\section{Introduction}\label{sec1}

Most real tasks involve more than one agent, and those agents closely interact with each other. In a diverse multi-agent world, figuring out who to cooperate with and who to protect oneself against is a fundamental challenge for any agent. 
In some artificial tasks, the identities of agents (i.e., competitors or cooperators) are public and fixed during the game. Notably, significant achievements have been made in developing machine learning models to handle such tasks, e.g., Go \cite{1,2,3}, Moba \cite{7,8}, Doudizhu (the most popular card game in China \cite{4,5,6}), and football \cite{9,10}. However, in real life, there are more tasks where the agents' identities are ambiguous, and they may even dynamically change over time. 
For example, the question that who is on my team, often arises on the first day at the kindergarten or the stock exchange. Besides, the drivers also need to identify the intentions of surrounding vehicles.  
This implies that the agents in the hidden-role environments have to decide to either compete or cooperate with others, and endowing the agent with the ability to accurately identify the intentions of other agents is a critical research problem in the multi-agent reinforcement learning (MARL) domain \cite{bucsoniu2010multi}.

MARL is a prevalent learning paradigm for stochastic games. There are three typical environments explored in MARL: completely cooperative, completely competitive, and mixed. In a completely cooperative setting, all agents share the same global reward, and they have to learn to cooperate to the maximum extent so that they can maximize their individual rewards. 
There is rich literature on dealing with cooperative MARL problems (e.g., credit assignments \cite{11,12,13,25,26} and communications \cite{14,15,23,24}).  For completely competitive tasks (e.g., zero-sum games), agents have the opposite goal, which is to get more rewards than others. Hence, they must learn to compete with and outmaneuver other agents. The challenges in this situation include creating a meaningful opponent via self-play \cite{17,18} and modeling the opponents \cite{16}. For mixed tasks, each agent is given (or develops) its own \textit{identity}, which compels it to either cooperate or compete with the other agents. Most research in this area assumes that the identities of these agents are known to all and remain fixed for the duration of the task \cite{7,8}. Hence, there has been scant research exploring games in which the identities of agents are ambiguous. Notably, this is a common case in most real-world scenarios. The core challenge in the problem lies in that the information about the agents’ identities is ambiguous and noisy. It is challenging to figure out the opponents and teammates only via the local imperfect information. Moreover, there may be deception in these hidden-role environments.

To solve the above challenges, we propose a novel identity detection reinforcement learning (IDRL) framework that dynamically identifies the agents’ ambiguous identities and selects the corresponding policies from the policy module for completing the tasks. In the IDRL framework, we have developed an identification module consisting of a relation network and a danger network. The relation network assigns a confidence value indicating the probability of the other agent being cooperative or competitive, and the danger network learns to generate a risk ratio representing the cost of mistaking the other agent’s intentions. The identification module detects the identity of the agent, so that each agent knows whom to cooperate with. Then, a policy module selects the policy to cooperate with teammates or compete with opponents respectively. Beyond that, we propose an intrinsic reward for seeking the trade-off between maximizing external rewards and reducing the risk of incorrect identification. Note that the proposed policy module is generic and compatible with any standard MARL method.

We evaluate the performance of IDRL in the \textit{Red-10} card-shedding game environment, where the identities of agents are ambiguous. Experiment results show that IDRL significantly outperforms other MARL methods, including mean-field multi-agent reinforcement learning (MFRL), Douzero (i.e., a reinforcement learning framework for the DouDizhu card game), and CQL. We also find that IDRL performs on par with the top human Red-10 players. In particular, ablation studies and additional experiments demonstrate the remarkable efficacy of the identification module and the danger network. 
Our contributions are threefold. First, we propose a novel MARL framework, IDRL, to learn policies with identification capability in the scene where agents’ identities are ambiguous. Such hidden-role environment is quite general, e.g., Avalon \cite{serrino2019finding} and Werewolf \cite{wang2018application}.
Second, we develop an identification module consisting of a relation network and a danger network to detect identities accurately. Third, we propose an intrinsic reward for policy learning which balances the need to maximize external rewards and the need to reduce risks.

In Section \ref{sec2}, we introduce the notations of this work. After that, we describe the details of the proposed framework IDRL in Section \ref{sec3}, which is composed of a relation network and a danger network.
 Next, we explain the proposed intrinsic rewards and the Monte Carlo-based \cite{27} policy module.
 In Section \ref{related}, we discuss the related works.
 In Section \ref{sec5}, we first provide the implementation details of IDRL, and then show the comparison experiment results.
 Furthermore, we analyze the effects of the different parts of the IDRL framework and provide interpretations and discussions. Finally, Section \ref{sec6} offers the conclusion of this work and points out some future directions.

\section{Background}\label{sec2}

We consider the MARL problem to be a Markov game \cite{22}, which is defined by a tuple $(N,S,A,P,R,\gamma)$. Here, $N=\{1,2,\ldots,n\}$ is the set of agents, $S$ is the finite joint state space, $\textbf{s}=\Pi_{i=1}^{n}s_i\ \ \forall \textbf{s} \in S$, and $s_i$ is the local state of agent $i$. The joint action space ($A =\Pi_{i=1}^{n}A_i$) is the union of all agents’ finite action spaces, $P:S\times A\times S\rightarrow \mathbb{R}$ is the transition probability, $r: S\times A\rightarrow \mathbb{R}$ is the reward function, $r_i(\textbf{s},\textbf{a}) = O_i(r(\textbf{s},\textbf{a}))$, for $ \forall \textbf{s}\in S, \forall \textbf{a} \in A$ is the reward of agent $i$, and $\gamma \in [0,1)$ is a discount factor. The goal of the agent $i$ is to learn the optimal policy, $u_i^*:S \rightarrow A_i$, and under the optimal policy $u_i^*$, the expected cumulative discounted reward, $V_i^\textbf{u}(\textbf{s})$ could be maximized,
\begin{equation}
  V_i^{\bm{u}}(\textbf{s})=\mathbb{E}_{\bm{a}^t=\bm{u}(\bm{s}^t),\bm{s}^{t+1}\sim P}[\sum_{t=0}^\infty \gamma^tr_i(\bm{s}^t,\bm{a}^t) \lvert \bm{s}^0=\bm{s}],
\end{equation}
where $\textbf{u}=(u_1,u_2,\ldots,u_n)$ is the product of all agents' policies, and ($\textbf{s}^t,\textbf{a}^t$) is the global state and action at time step $t \in \mathbb{N}$. The optimal policy of agent $i$ can be obtained by maximizing the $Q$ action-value function as follows:
\begin{equation*}
Q^{\textbf{u}}(\textbf{s},\textbf{a})=\mathbb{E}_{\textbf{a}^t=\textbf{u}(\textbf{s}^t),\textbf{s}^t\sim P}[r_i(\textbf{s}^0,\textbf{a}^0) + \sum_{t=1}^\infty \gamma^tr_i(\textbf{s}^t,\textbf{a}^t)\lvert \textbf{s}^0=\textbf{s},\textbf{a}^0=\textbf{a}]
\end{equation*}
and $u_i^*(s)=\arg\max_{a_i}max_{\textbf{a}_{-i}}Q^*(\textbf{s},\textbf{a})$, $\textbf{a}_{-i}$ is the joint action set of all agents except agent $i$, for whom $Q^*(\textbf{s},\textbf{a})=\max_{\textbf{u}}Q^{\textbf{u}}(\textbf{s},\textbf{a})$.

$S_{re},S_{dan}$ are two sub-sets of $S$. $S_{re}$ is the joint state of the relation network, and $S_{dan}$ is the joint state of the danger network. $\textbf{s}_{re}^t =\Pi_{i=1}^{n}s_{re}^{t_i}\ \ \forall \textbf{s}_{re}^t \in S_{re}$, $s_{re}^{t_i}$ is the the state input to agent $i$’s relation network at time step $t$, and $\textbf{s}_{dan}^t =\Pi_{i=1}^{n}s_{dan}^{t_i}\ \ \forall \textbf{s}_{dan}^t \in S_{dan}$, $s_{dan}^{t_i}$ is the the state input of agent $i$'s danger network at time step $t$.

\section{Method}\label{sec3}

To solve the challenging multi-agent games with ambiguous identities, we introduce a novel identity detection reinforcement learning (IDRL) framework, which is composed of an identification module and a policy module.
  In subsection \ref{sec3.2}, the relation and danger networks in the identification module are described. Beyond  those two networks, we further develop an intrinsic reward function to facilitate the optimization of the proposed networks. 
  In subsection \ref{sec3.3}, the Monte Carlo-based policy module is discussed, which is used to do the decision-making after identification.
  In subsection \ref{sec3.4}, we provide the parallel training method used in IDRL training.

  \begin{figure*}[ht]
\centering
\includegraphics[width=\textwidth]{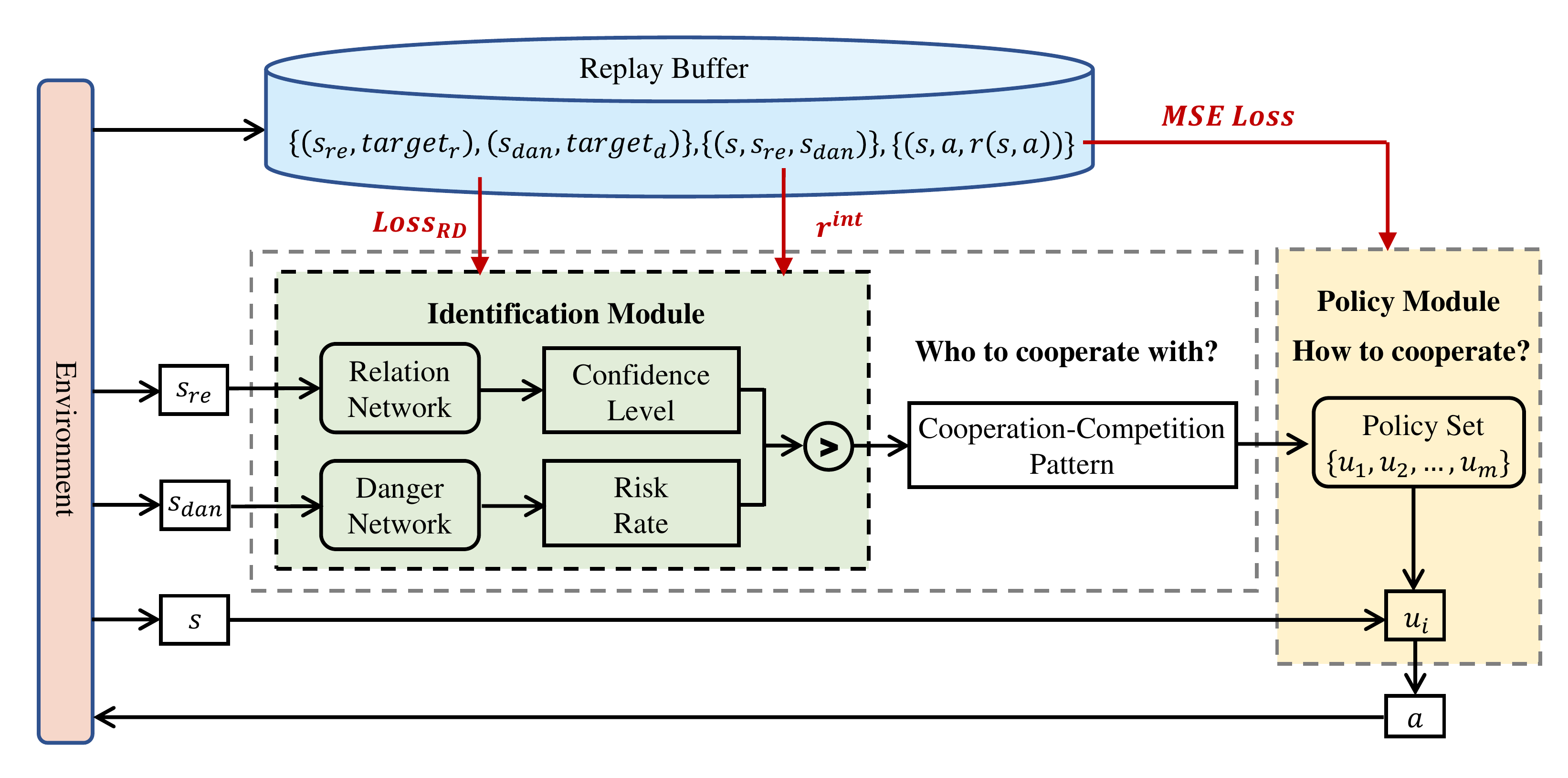}
\caption{Overview of the identity detection reinforcement learning (IDRL) framework. ``$>$’’ indicates the comparison of confidence level and risk rate. There are $m$ cooperation-competition pattern types, and the red arrows represent the flow of gradients. MSE: mean squared error.}
\label{fig1}
\end{figure*}

\subsection{The IDRL framework}
The core concept of the IDRL framework is to transform a setting with ambiguous agent identities into one with less ambiguous identities. That is, agents are empowered to intuitively infer the identity of a cooperating agent and then act upon the assumption. 

Fig. \ref{fig1} visualizes the IDRL framework, which consists of two modules: identification and policy. The input to the IDRL framework includes the encoded local observation (and legal action set). The output is the chosen policy (action set). The identification module processes information first; then, the policy module, which is pretrained with appropriate action sets, generates operational rule sets. Recall that the identification module comprises relation and danger networks. The relation network assigns each agent a confidence level reflecting the perceived probability of the agent under consideration being a teammate. The danger network then generates a risk ratio by perceiving the task at hand and analyzing the potential loss if a mistake in identification judgment is made. The confidence level and risk ratio are then combined to select a corresponding policy from the policy module. Then, the agent acts upon the selected policy.

Here, we describe the notation of the IDRL framework. $R_{\theta}(s_{re}^{t_i})$ is the relation network with parameter $\theta$, $s_{re}^{t_i}$ is the state input of the relation network of agent $i$ at time step  $t$. The output of the network is a vector, $\textbf{c} \in \mathbb{R}^{n-1}$, and each value inside the vector is a confidence level corresponding to another agent. $D_{\alpha}(s_{dan}^{t_i})$ is the danger network with parameter $\alpha$, and the state input to the relation network of agent $i$ at time step $t$ is  $s_{dan}^{t_i}$. The output of the network is the risk ratio, $\textbf{d} \in \mathbb{R}$. When the agents’ identities are ambiguous, the action $a_i^t$ taken by agent $i$ in state $s_i^t$ at time step $t$ is specified by the policy, $\pi_{\beta_i}(s_i^t,R_{\theta}(s_{re}^{t_i}),D_\alpha(s_{dan}^{t_i}))$, where $\beta_i$ is agent $i$'s policy parameter. The policy combines the state, confidence level, and risk ratio to decide upon an action that should result in the largest cumulative reward based on the given state.

\subsection{Identification Module}\label{sec3.2}
Here, we discuss the relation and danger networks and elaborate upon their training method and loss function. Finally, we propose a novel intrinsic reward function to facilitate the learning of both networks.

\subsubsection{Relation Network}
The relation network predicts the confidence levels of other agents being a teammate. It is the primary part of the identification module. The network’s input is the state information, including historical agent actions and local observations. The output is a \begin{math}n-1\end{math}-element vector, and each value in the vector reflects the confidence level of an agent.

The relation network is important for games with hidden or ambiguous identities and imperfect information environments, such as Red-10, Avalon, and Werewolf.
Naturally, the information acquired by each agent quickly becomes asymmetric, and the agent’s perceived identities of others differ from agent to agent. The relation network is designed for such an asymmetry information setting, which observes the behaviors of other agents
and predicts the probability of the agent being a teammate. 

\subsubsection{Danger network}
In an imperfect information environment, an agent needs to estimate the importance of action taken at a time step in terms of the potential reward, and compare it to the risk of trusting another entity with ambiguous identities. The danger network derives this risk ratio from the input of the state information, including the historical actions and local observations of other agents. 
A larger risk ratio indicates more danger of making an incorrect guess.

The core concept of the danger network is to cooperate as possible when the risk is low and to compete as much as possible when the risk is high. At the beginning of a new round, it is natural to find that less risk exists; hence, agents cooperate with others at a high rate, which results in choosing more reasonable and beneficial actions to get more reward, not considering the actions just for suppression. In contrast, over time, the risk continues to grow, and it is risky to cooperate with an agent whose identity is much ambiguous, so a better strategy is to compete with such agents to avoid the opponents winning finally.

 Next, we present how the identification module uses the outputs of the relation and danger networks to identify a teammate. At each time step, the relation network predicts a confidence level vector for every agent, and the danger network outputs a risk ratio. Then, each agent, $i$, compares the confidence level of another agent with the risk ratio. If the confidence level of another agent, $j$ $(j\in \mathbb{N}\ and\ j \neq i)$, is bigger than the respective risk ratio, agent $i$ tends to cooperate with agent $j$. Otherwise, competition is warranted. During the same time step, each agent finally forms the intention to either cooperate or compete with all other agents. After the identification, the corresponding cooperation--competition policy can be used just as in the setting where the identities of the agents are public.

\subsubsection{Loss Function}
\label{network}
\begin{figure}[htbp]
\centering
\includegraphics[width=1.0\textwidth]{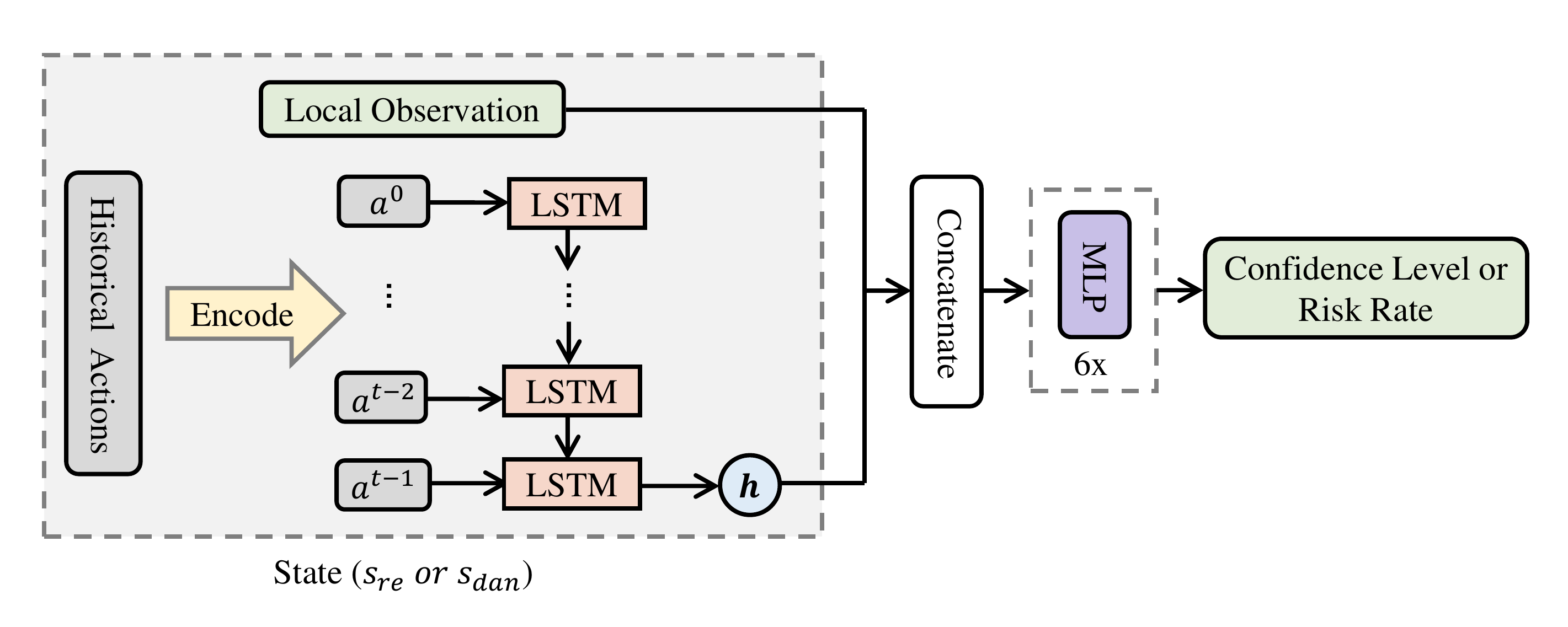}
\caption{Architecture of the relation and danger networks. LSTM: long short-term memory; MLP: multilayer perceptron.}
\label{fig:2}
\end{figure}

The relation network and the danger network use the same network architecture, as shown in Fig. \ref{fig:2}. A long short-term memory (LSTM) is used to encode the historical actions, and the output is concatenated with local observations as state information. A six-layer multilayer perceptron (MLP) is then used to predict the confidence level or risk ratio.

During training, as the global information is made available, the identity of each agent becomes known, and we use the ground-truth relationship vector, $target_r^{i}$, as the target output of the relation network for agent $i$. $target_r^i[j]=1$ or $target_r^i[j]=0$ signify if the agent $j$ is a teammate or an opponent, respectively. At the end of a round, the task completes with time \begin{math}T\end{math}, and \begin{math}target_d^t=\frac{t}{T}\end{math} is set as the target of the output of the danger network at time step \begin{math}t\end{math}. The loss function of the relation and danger networks is shown in Eq.(\ref{equal1}), where we first use the mean-square-error (MSE) loss function (MSELoss) to update the networks via gradient descent operations. Then, we use the intrinsic reward of Eq.(\ref{equal2}) to further update the two networks via gradient ascent operations, which leads to a trade-off assessment between the need to maximize the external reward and the accuracy of identification.

\begin{equation*}
    Loss_{RD}=\sum_{t=0}^T\sum_{i=0}^n\operatorname{MSELoss}(R_\theta(s_{re}^{t_i})-target_r^{i})
\end{equation*}
\begin{equation}
    \label{equal1}
    +\sum_{t=0}^T\sum_{i=0}^n\operatorname{MSELoss}(D_\alpha(s_{dan}^{t_i})-\operatorname{target}_d^{t})
\end{equation}

\subsubsection{Intrinsic Rewards}
Considering that the goal of identification is to maximize the expected cumulative external reward, we develop the intrinsic reward as Eq.(\ref{equal2}) to further update the relation network and the danger network via gradient ascent operation, so that the agent can obtain a larger reward if the accuracy of identification is satisfactory.
\begin{equation}
     r^{int} = \sum_{t=0}^{T}\sum_{i=0}^{n}Q_i\left(s_i^t, \pi_{\beta_i}\left(s_i^t, R_\theta\left(s_{re}^{t_i}\right),D_\alpha\left(s_{dan}^{t_i}\right)\right)\right)-\lambda\operatorname{Loss} _{RD},
    \label{equal2}
\end{equation}
where the $\operatorname{Loss} _{R D}$ is the identification loss in Eq.(\ref{equal1}). $Q_i$ is the state-action value function of agent $i$, and the action is decided by policy $\pi_{\beta_i}$, based on both the local state and the outputs of the relation network and danger network. $\lambda$ is a hyper-parameter used to balance external rewards and identification accuracy.

\subsection{Policy Module}\label{sec3.3}
Based on the identification results, one agent regards others as teammates or opponents, and the policy module provides a range of actions that can be applied during their cooperative or competitive interactions. As the diversity of the identification result, we prepare sets of policies in the policy module for different cooperation--competition patterns to cooperate or compete with different agents. Our policy module is generic and ready to be applied with existing MARL methods such as \cite{21,28,29,30,31,32}. In this work, we use the deep Monte Carlo method to obtain the policy sets. For more complicated tasks, we can apply other advanced MARL methods in the policy module.

\begin{figure}[htbp]
\centerline{\includegraphics[width=0.95\textwidth]{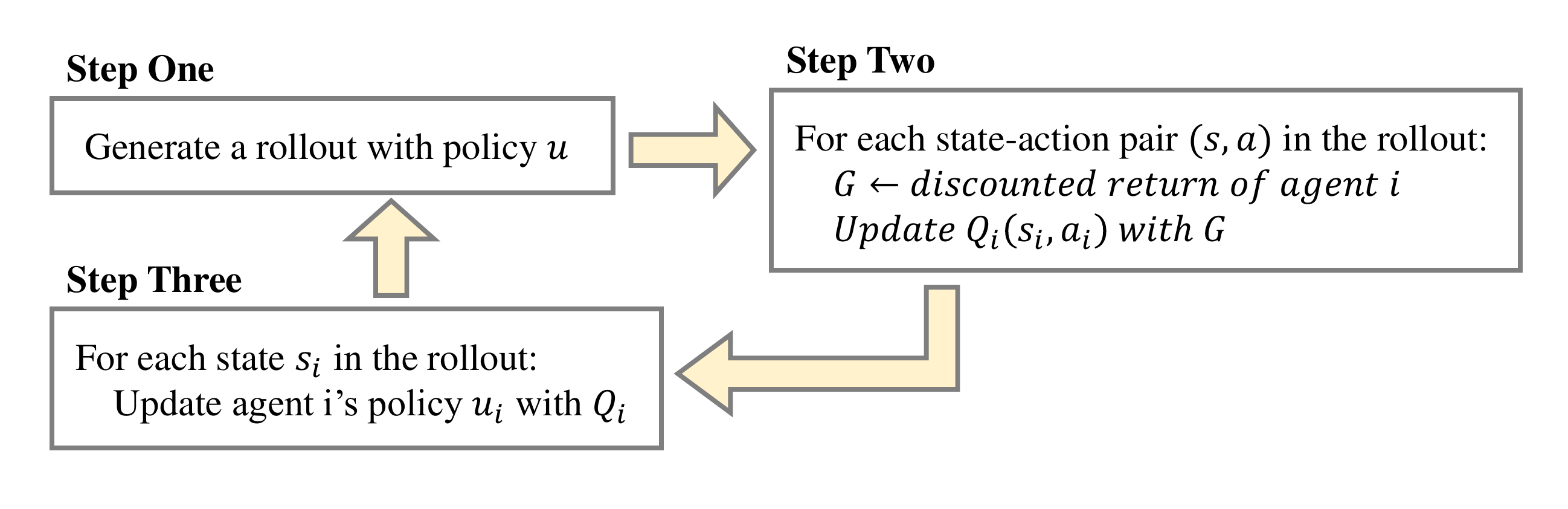}}
\caption{Steps of the Monte-Carlo method.}
\label{fig3}
\end{figure}

The Monte Carlo method is a traditional reinforcement learning method \cite{27} designed for tasks that require rounds of activities. Fig. \ref{fig3} shows the three main  iterative steps toward estimating a \begin{math}Q_i\end{math} action-value function for agent \begin{math}i\end{math} to optimize its policy, \begin{math}u_i\end{math}. In step one, we employ the $\epsilon$-greedy strategy (Eq.(\ref{equal3})) to balance exploration and exploitation. In step two, the action-value function directly approximates the actual cumulative discounted reward. Then, in step three, we update the policy as \begin{math}u_i(s_i)=\arg\max\limits_{a_i}{Q_i(s_i,a_i)}\end{math}. In deep Monte Carlo, we create a neural network to approximate the action-value function and MSE to update the network. 


\begin{equation}
{a_i} = \begin{cases}
choose\ action\ randomly,&{with\ probability\ \epsilon} \\ 
{u_i(s_i),}&{with\ probability\ 1-\epsilon}
\end{cases}
\label{equal3}
\end{equation}

\subsection{Parallel Training}\label{sec3.4}
We divide IDRL model training into three parts. The policy module is trained to obtain the policy set first. Then, the relation and danger networks are trained together. Finally, all the networks are updated with the proposed intrinsic rewards.

Expecting to get more interacting data, and, inspired by \cite{33}, we use a parallel network training method with actors and learners. Each actor process constantly plays the task using the local network and generates new behaviors, the local network is then updated with information from the global network in the learner process at the beginning of each round, and experiences are saved to the shared buffer at the end of each round. The learner process maintains the global network and constantly updates using the data in the shared buffer. The details of the actors’ and learners’ Q action-value network updates are provided in Algorithms \ref{algo1} and \ref{algo2}, respectively. Note that in every GPU, we can run several parallel actors per learner.

\floatname{algorithm}{Algorithm}
\begin{algorithm}[h]
\caption{: Actor}
\begin{algorithmic}[1]
\Require {\bf shared buffer: $B$, buffer size: $BS$, discount factor: $\gamma$}
\item initialize the local network: \begin{math}Q\end{math}, initialize local buffer: \begin{math}D\end{math}
\For{i=1, 2, 3 \ldots}
\State update \begin{math}Q\end{math} with the global network \begin{math}Q^g\end{math} in learners
\For{t=0, 1, 2, \ldots, T}\ \ \ \ \ \ \ \ (taking action)
\State 
$a^t\gets \left\{
\begin{array}{lll}
\arg\max\limits_{a}{Q(s,a)}       &      & {with\ prob\ 1-\epsilon}\\
random\ action     &      & {with\ prob\ \epsilon}\\
\end{array} \right. $
\State execute the action $a^t$; get reward $r^t$
\State put data \begin{math}(s^t,a^t,r^t)\end{math} into local buffer \begin{math}D\end{math}
\EndFor
\For{t=T-1, T-2, \ldots, 0}\ (calculating discounted returns)
\State $G^t\gets r^t+\gamma G^{t+1}$
\EndFor

\If{$D.length\ge BS$}\ \  (saving the data to $B$)
    \State request an empty $B$
    \State move a batch of \begin{math}(s^t,a^t,r^t)\end{math} of size $BS$ from $D$ to $B$
    \EndIf
\EndFor
\end{algorithmic}
\label{algo1}
\end{algorithm}

\begin{algorithm}[htbp]
\caption{:Learner} 
\begin{algorithmic}[1]
\Require  {\bf shared buffer: $B$, batch size: $M$, learning rate: $\psi$ }
\item initialize the global network: $Q^g$
\For{i=1, 2, 3, \ldots}
\If{the data size in $B$ $\ge M$} 
    \State pop $M$ data from $B$
    \State use MSE loss and learning rate $\psi$ to update the $Q^g$ network
    \EndIf
\EndFor
\end{algorithmic}
\label{algo2}
\end{algorithm}

\section{Related Work}\label{related}
This section reviews the related works about multi-agent reinforcement learning and ad hoc teamwork.

\noindent\textbf{Multi-agent Reinforcement Learning (MARL)} investigates the learning problem containing multiple autonomous agents. Typically, the learning scenarios in MARL could be divided into three categories: fully cooperative games, fully competitive games, and mixed games. In fully cooperative learning environments, existing research focuses on the problems of multi-agent communication \cite{14,15,23,24}, collaborative exploration \cite{wang2019influence, liu2021cooperative, viseras2016decentralized}, and credit assignment \cite{11,12,13,25,26}. 
In fully competitive games, e.g. zero-sum games, the agents aim to obtain more rewards than others, so that their relationships are competitive. To conquer other agents, previous works propose to learn competitive policies with self-play \cite{17,18} or opponent modeling \cite{16, he2016opponent}. In mixed games, each agent either cooperates or competes with others based on its inherent property (identity). Most previous works in this domain assume that the identities of these agents are public and remain fixed during the game \cite{7,8}. However, a more general case is that the agents do not know who is a friend and who is a foe, for example, on the first day of kindergarten or the stock exchange. There is seldom research work dealing with this challenging problem of detecting the identities of other agents in mixed games. This work tries to solve this challenge by learning an identification module composed of a relation network and a danger network.

\noindent\textbf{Ad Hoc Teamwork} aims to improve the adaptation ability of the agents to collaborate with diverse and unknown teammates \cite{stone2010ad, mirsky2022survey}, which also involves inferring the types of teammates. Specifically, PLASTIC \cite{barrett2015cooperating} computes the Bayesian posteriors over all the types of teammates. ConvCPD \cite{ravula2019ad} allows the changing of teammate types, which introduces a mechanism to detect the change point of the current type. AATEAM \cite{chen2020aateam} develops an attention-based structure to infer teammate types from the state histories. ODITS \cite{gu2021online} proposes a multimodal representation framework to encode teamwork situations.
The previous ad hoc teamwork methods mostly work in a fully collaborative setting and focus on fast adaptation to varying teammates. In contrast, our work detects relationships among the agents in hidden-role scenarios with mixed collaboration and competition.

\section{Experiments}\label{sec5}
 Based on the toolkit for reinforcement learning in card games \cite{19}, we develop a Red-10 game environment and provide corresponding evaluation and visualization tools. In this section, we first introduce the rules of Red-10, and then provide the implementation details of IDRL. After that, we show the experiment results. The experiments are designed to answer the following questions.
\begin{enumerate}
    \item How does IDRL compare with existing SOTA MARL methods? (subsection \ref{sec5.1})
    \item Can the relation network accurately identify the cooperation and competition intention of the agents? How does the  inference capability of the relation network compare with that of a human? (subsection \ref{sec5.2})
    \item Does the danger network reduce the risk of acting upon no-perfect identifications generated by the relation network?  (subsection \ref{sec5.3})
    \item Does the identification module enhance the capacity of decision-making? (subsection \ref{sec5.4}) 
\end{enumerate}

\noindent{\bf Evaluation Metrics:} To evaluate our model performance, we followed \cite{34} by comparing the win rate of our algorithm with that of other methods. To compare the performance of algorithm $X$ with algorithm $Y$, $X$ controls agent 0 (the agent whose id is 0), and $Y$ controls the other three to play 50000 decks in the Red-10 game environment. To reduce the variance, $X$ and $Y$ switched agents and played another 50000 decks. The entire process was repeated four times; then, we calculated the normalized win rates of $X$ and $Y$.

\subsection{Rules of Red-10}
Red-10 is a variation of Doudizhu and has recently grown in popularity. Like most card-shedding games, the goal of players is to play all their cards before the others do. In Red-10, four players use a standard card deck, shuffled, and the full deck is dealt so that each player has an equal number of cards. Players are then divided into ``Landlord’’ and ``Peasant’’ teams, where the players with the red-suited (hearts or diamonds) ``10’’ cards (red 10 card) are on the Landlord team. Note that there are only two red 10 cards in a standard deck, and each player either has a red 10 and does not know which of the other three has the other or lacks a red 10 and does not know which two or one of the other three players have them. Hence, each agent lacks knowledge of the identity of their teammate. The four players then compete by taking turns playing a card and observing the others’ actions in order to build a framework for identifying their teammates, because teammates can cooperate with one another while competing with members of the other team. The game finishes when a player runs out of cards (wins). Hence, that player’s team also wins. Additional rules are provided in Appendix A.

Red-10 provides an imperfect information environment in which agents can only make local observations of their own cards while compiling knowledge about the actions of the other agents. The interplay is further complicated by behavioral ``noise’’ (e.g., bluffing) and the large action space in which there are about 27,400 possible combinations. Various subsets of these combinations are legal action sets for different hands.

\noindent{\bf Agent Settings:} Four agents participate, and each is with a unique ID in ${0,1,2,3}$. They formed a circle as illustrated in Fig. \ref{fig7}. The agents played cards in turn counterclockwise recurrently around the circle. For convenience, from an agent’s perspective, we deemed the player on its left as ``up’’ the opposite player as ``front’’ and the player on the right as ``down’’. For example, for agent 0, the up player is agent 3, the front player is agent 2 and the down player is agent 1. 

\begin{figure}[htbp]
\centering
    \subfigure[Encoding of the card combination: We encode every type of card combination to a \begin{math}4\times13\end{math} one-hot matrix. Each column of the matrix represents a type of face value card in the vector below, and the sum of the column is the amount of the face value card in the combination. The upper cards combination is encoded into the matrix middle.]{
    \begin{minipage}[t]{0.45\textwidth}
        \centering
        \includegraphics[width=1\textwidth]{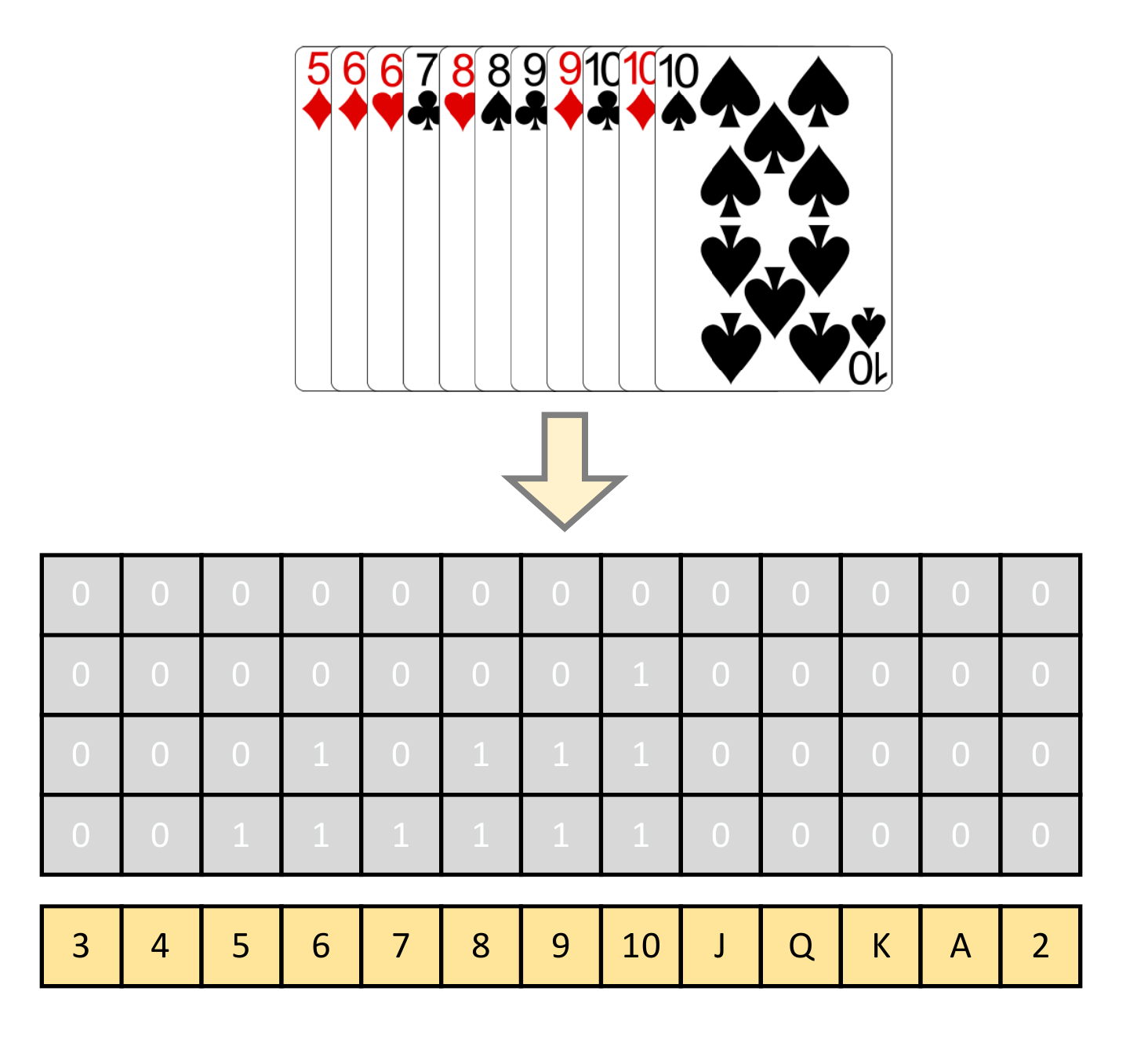}
        \label{fig4:a}
    \end{minipage}}
    \subfigure[Encoding of the card suit: We encode every type of suit combination of card 10 to a \begin{math}1\times4\end{math} one-hot vector. Each number of the vector represents a type of suit, including Heart, Diamond, Club, and Spade. The suit of card 10 is in the combination if the corresponding number in vector is one. The upper suit combination is encoded into the vector middle. ]{
    \begin{minipage}[t]{0.45\textwidth}
        \centering
        \includegraphics[width=0.55\textwidth]{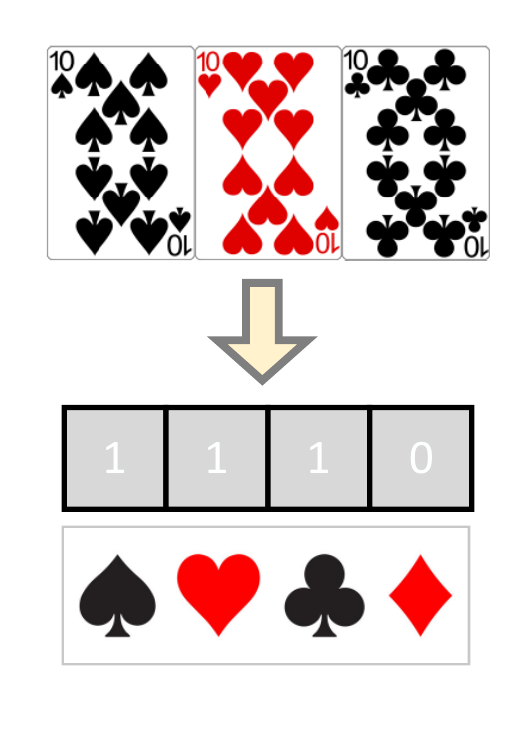}
        \label{fig4:b}
        \end{minipage}}
\caption{Card representations.}
\label{fig4}
\end{figure}

\subsection{Implementation Details}\label{sec4}
In this section, we describe the implementation details of IDRL in the Red-10 game.

\subsubsection{Observations and Network Architectures}
Based on the card representation in \cite{6}, in the Red-10 game, the observations are encoded by the way shown in Fig. \ref{fig4}. We encode the cards in one hand, the union of cards in the other hand, and the action into \begin{math}4\times13\end{math} matrices. We encode the information about the color of card 10 into \begin{math}1\times4\end{math} one-hot vectors.

\begin{figure}[htbp]
\centerline{\includegraphics[width=1.\textwidth]{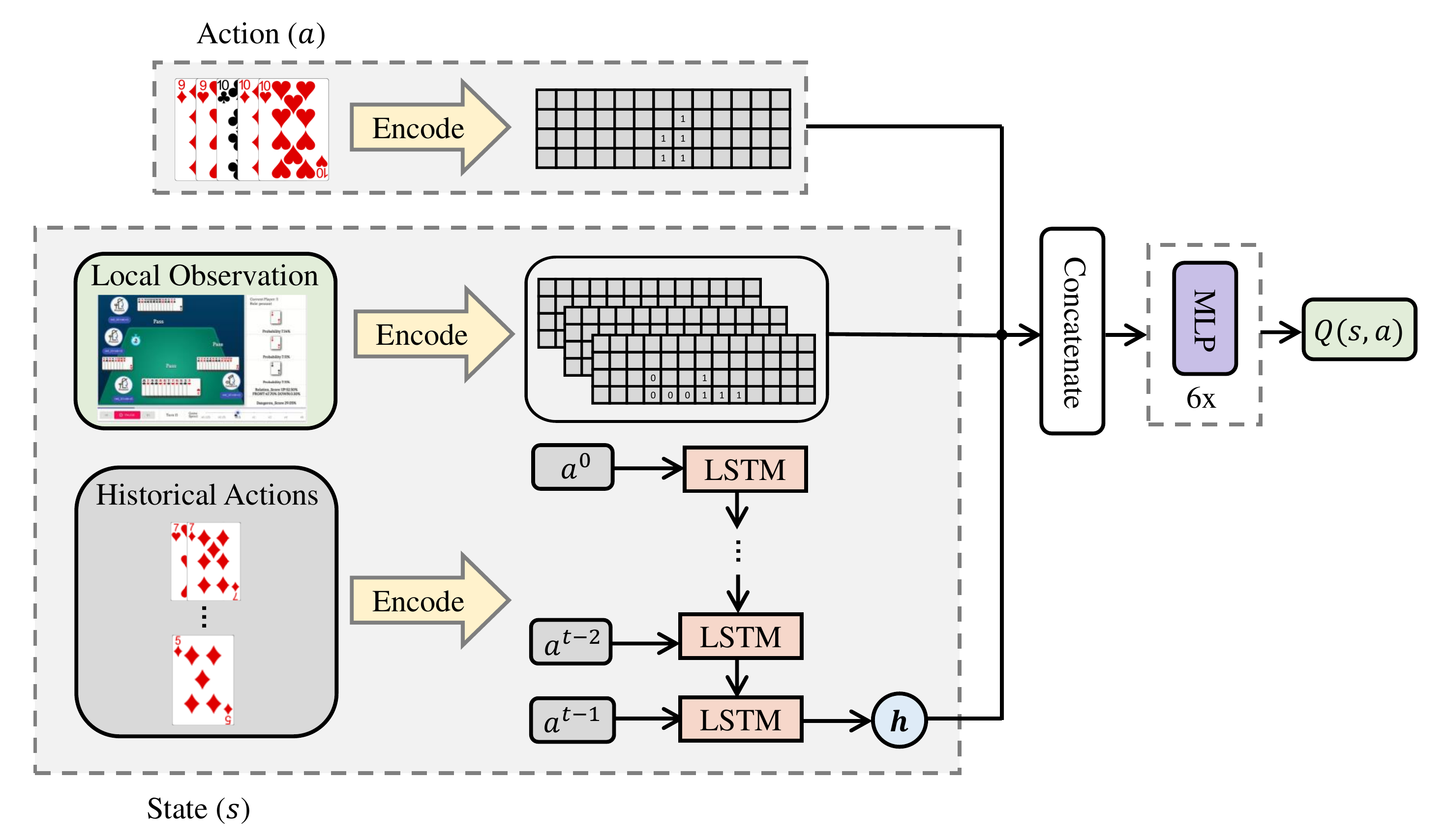}}
\caption{Q network in the policy module. LSTM: long short-term memory; MLP: multilayer perceptron.}
\label{fig5}
\end{figure}

Fig. \ref{fig5} shows the architecture of the \begin{math}Q\end{math} network in the policy module. The input includes the state and the action. The state consists of the encoded local observation and historical card-playing information using the LSTM encoder. We concatenate the encoded state and action as input to the multi-layer perceptrons. The output is the Q value. The details of the input data could be found in Appendix B.

The architectures of the relation network and the danger network are described in subsection \ref{network}. As the information about the color of card 10 is vital for identification, we add several encoded suit sets of card 10 to the input except for the encoded local observation and historical actions. Details of the data input into the relation and danger networks are found in Appendix B.

\subsubsection{Training Details}
\begin{figure}[htp]
\centering
    \subfigure[Pattern 1100 : Two landlords and two peasants; agents on same team are adjacent]{
    \begin{minipage}[t]{0.2\textwidth}
        \centering
        \includegraphics[width=1\textwidth]{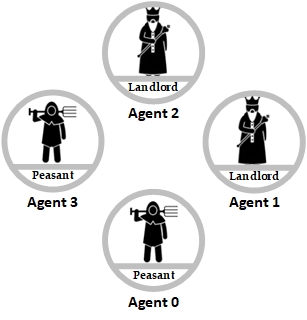}
    \end{minipage}}
    \subfigure[Pattern 1010 : Two landlords and two peasants; agents on the same team are opposite]{
    \begin{minipage}[t]{0.2\textwidth}
        \centering
        \includegraphics[width=1\textwidth]{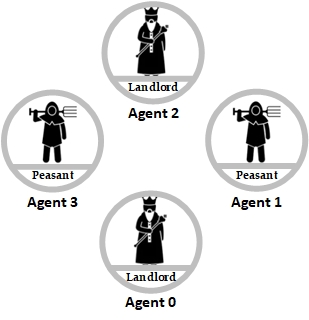}
        \end{minipage}}
    \subfigure[Pattern 1000 : Three agents in a team]{
    \begin{minipage}[t]{0.2\textwidth}
        \centering
        \includegraphics[width=1\textwidth]{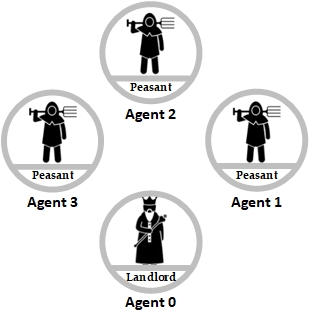}
    \end{minipage}}
    \subfigure[Pattern 0000 : Four agents in a team]{
    \begin{minipage}[t]{0.2\textwidth}
        \centering
        \includegraphics[width=1\textwidth]{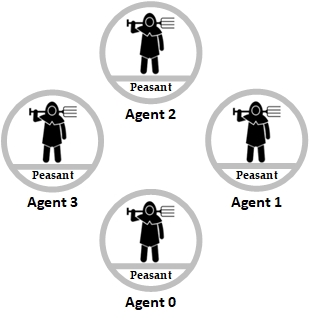}
    \end{minipage}}
\caption{Different agent patterns.}
\label{fig7}
\end{figure}

As shown in Fig. \ref{fig7}, there are four cooperation--competition patterns, including three types of agent location distributions in the Red-10 game and the situation the agent wants to cooperate with the other three agents (i.e., the agent’s cooperation decisions). The four cooperation--competition patterns are ``1100’’, ``1010’’, ``1000’’ and ``0000’’ respectively, where the ``1’’ and ``0’’ represent ``Landlord’’ and ``Peasant’’ respectively. Thus, we must train four corresponding policy sets in the policy module. For example, with the distribution shown in Fig. \ref{fig7}(b), there are two landlords and two peasants, and teammates sit opposite one another (i.e., ``1010’’). The policy for this cooperation--competition pattern is designed to help an agent cooperate with the teammate agent across the table while competing with the two to the left and right. Then, the relation and danger networks are trained together. Finally, the intrinsic reward is used to further update the identification module.

\subsection{Experiment Results}\label{sec5.1}
We compared IDRL to the following state-of-the-art MARL methods.
\begin{itemize}
    \item {\bf Douzero \cite{6}: }Douzero is one of the most powerful MARL models designed to handle the classic three-person Doudizhu game, and it has demonstrated par performances with the top human players. Because Red-10 game is a variation of Doudizhu, the action, and state spaces overlap. Therefore, it is appropriate to compare IDRL with Douzero. 
    \item {\bf Combinational Q-learning (CQL) \cite{20}: }This is another reliable three-person Doudizhu-playing system that uses the MARL method. A two-stage network is employed to reduce the action space and leverage order-invariant max-pooling operations to extract relationships between primitive actions. We modified its code to adapt it to Red-10.
    \item {\bf Mean Filed Reinforcement Learning (MFRL) \cite{21}: } MFRL is the baseline MARL method used to solve mixed cooperative and competitive tasks. It approximates interactions within a population of agents via single agent-to-agent interactions and the average effects from neighboring agents or the overall population. We trained the MFRL for optimal policy performance using the mean-field Q-learning algorithm.
    \item {\bf Rule-based RLcard \cite{19} and Random: }These models determine card-playing tactics using naive fixed rule sets or sampling the legal action space uniformly.
\end{itemize}
Table \ref{table1} shows the normalized winning probability of IDRL against all baselines. IDRL’s win rate was the highest; it achieved the best performance overall and decided card playing appropriately in the Red-10 game environment, where the agents’ identities are ambiguous. Comparing with the rule-based methods, IDRL got the dominant performance, demonstrating that adopting reinforcement learning in IDRL is effective. Even though the policy module relied on the traditional Monte-Carlo method, IDRL outperformed the baseline MARL, CQL, Douzero, and MFRL owing to the benefits of its identification module.

Considering the MFRL method got the best performance in baseline, we further compared IDRL with MFRL method in three Red-10 cooperation--competition patterns. We set the IDRL and MFRL as landlord and peasant, respectively. Later, they switched. IDRL’s win rates are summarized in Table \ref{table1.2}, where it clearly outperformed MFRL in all patterns, except when IDRL was the landlord in patterns “1010” and “1000”. In the first case, IDRL was at a disadvantage because MFRL went first. In the second, IDRL had no teammate. In both cases, the identification module provided no advantages, because they can't or have no person to cooperate with. And also in pattern “1000”, the MFRL agent as the landlord receives a far lower win rate than the IDRL as the landlord, which further demonstrates the powerfulness of IDRL. Comparing the win rate of IDRL as a peasant with IDRL as a landlord, we find the landlord identity limits the performance of IDRL: for all patterns, the win rate of IDRL as a landlord is lower than as a peasant, because some agents in the Landlord team get more specific information about red card 10 than agents in the Peasant team, what results in the less effectiveness of the identification module work in Landlord team, and the win rate descend. It further demonstrates the promotion of the identification module to the IDRL.

\begin{table}[h] 
\setlength{\abovecaptionskip}{0.05cm} 
\centering
\caption{IDRL Win Rate.} 
{
\begin{tabular}{l| r r r r r }
	\hline
	     \textit{Method} &\textit{MFRL}&\textit{CQL}&\textit{Douzero}&\textit{Random}&\textit{RLcard} \\ \hline
	     \textbf{\textit{Win Rate(\%)}}&54.8&65.0&71.9&72.1&85.2\\ \hline
	
\end{tabular}
}
\label{table1}
\end{table}

\begin{table}[h] 
\setlength{\abovecaptionskip}{0.05cm} 
\centering
\caption{IDRL vs. MFRL with different cooperation-competition patterns. The table shows IDRL’s win rates versus MFRL (landlord and peasant) in three Red-10 gaming patterns (see Figs. \ref{fig7}(a), \ref{fig7}(b), and \ref{fig7}(c)).} 
{
\begin{tabular}{l r r r}
	\hline
	     \multirow{3}{*}{Identity}&\multicolumn{3}{c}{pattern}\\ \cmidrule(r){2-4}
	                          &\textbf{1100}&\textbf{1010}&\textbf{1000}\\ 
	                          &\textit{Win Rate(\%)}&\textit{Win Rate(\%)}&\textit{Win Rate(\%)}\\  \hline
	                   \textit{Overall}&54.84&47.83&64.02 \\
	                   \textit{Peasant}&57.73&52.41&87.71 \\
	                   \textit{Landlord}&51.95&43.25&40.33 \\

\end{tabular}
}
\label{table1.2}
\end{table}

\begin{table}[htbp] 
\setlength{\abovecaptionskip}{0.05cm} 
\centering
\caption{Confidence level.} 
\begin{tabular}{l| r r r}
	\hline
	      Agent Location&\textit{up}&\textit{front}&\textit{down} \\ \hline
	        \textbf{\textit{Confidence Level(\%)}}&0.456&0.451&0.029\\ \hline
\end{tabular}
\label{table2}
\end{table}

\subsection{Analysis of the Relation Network}\label{sec5.2}
In this section, we analyze the performance of the relation network in different rounds, which were classified based on whether agents received explicit information about the location of a red 10 card. If so, the round was labeled 
``partially public identities’’; otherwise, it was labeled 
``partially ambiguous identities’’.
\subsubsection{Partially Public Identities}
\begin{figure*}[htbp]
    \subfigure[agent 0]{
    \begin{minipage}[t]{0.23\textwidth}
        \centering
        \includegraphics[width=1\textwidth]{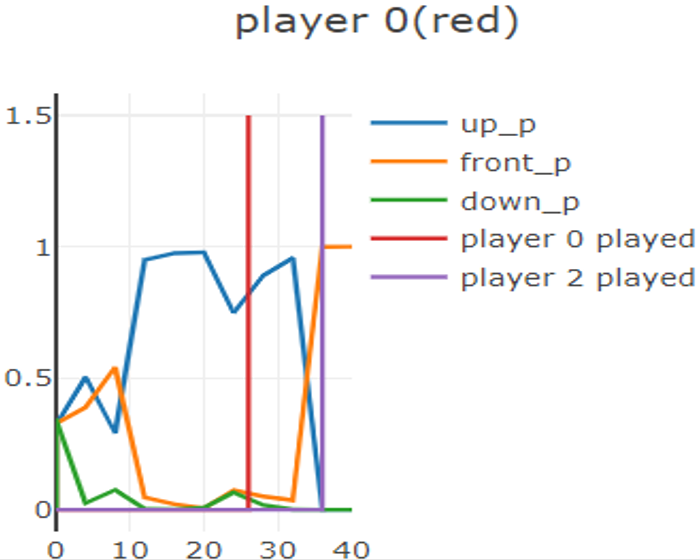}
    \end{minipage}}
    \subfigure[agent 1]{
    \begin{minipage}[t]{0.23\textwidth}
        \centering
        \includegraphics[width=1\textwidth]{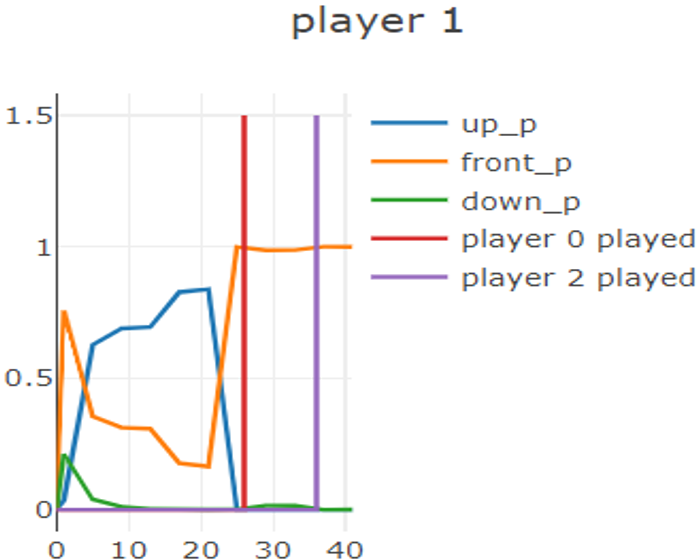}
        \end{minipage}}
    \subfigure[agent 2 ]{
    \begin{minipage}[t]{0.23\textwidth}
        \centering
        \includegraphics[width=1\textwidth]{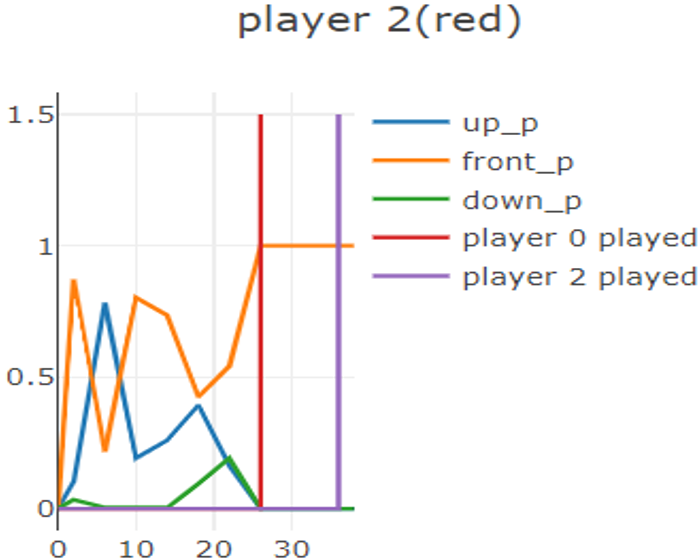}
    \end{minipage}}
    \subfigure[agent 3]{
    \begin{minipage}[t]{0.23\textwidth}
        \centering
        \includegraphics[width=1\textwidth]{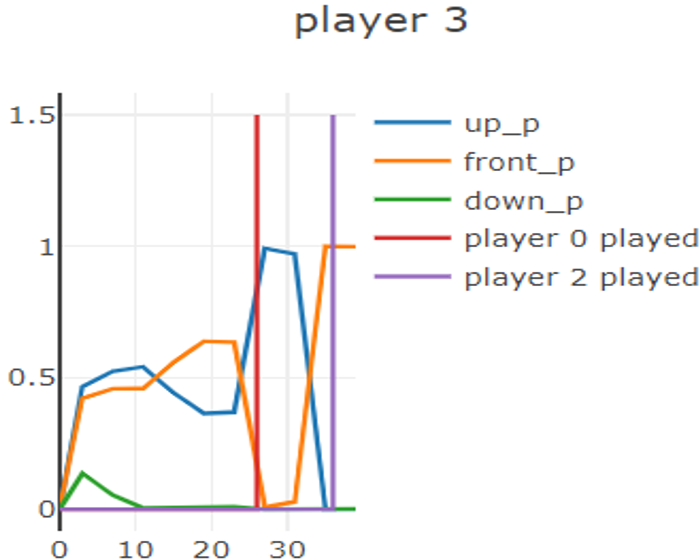}
    \end{minipage}}
\caption{Agents 0 and 2 hold a red 10 card, making them the Landlord team. Each subfigure shows the curve of an agent’s confidence level during a round. The title of a subfigure is the agent’s ID, and there is a tag as ‘(red)’ behind the ID if the agent ever holds a red 10 card. In a subfigure, the blue, orange, and green lines represent the confidence-level curves of the agent in question to the upper player (up\_p), the front player (front\_p), and the down player (down\_p), respectively. The vertical signifies the time step a landlord agent plays the red 10 card. The following confidence-level curves are also with partially public identities.}
\label{fig8}
\end{figure*}

This occurs when both red 10 cards are known by at least one agent or some agent plays the red 10 card; hence, the relation networks of some agents learn some or all of the team assignments. The following situations indicate this status:
\begin{enumerate}
    \item When two red 10 cards belong to a single agent (which can only occur during the first round), the agent is clearly aware that it has no teammate. Accordingly, its confidence level regarding the probability that the other three agents are teammates rapidly approaches zero. To verify this, IDRL played 1,000 decks with both red 10 cards assigned to a single agent each time. We then summarized the agent's mean confidence levels of other three agents. The results shown in Table \ref{table2} validate this conclusion.
    \item During any game (deck), if an agent (agent 0) plays a red 10 card, that agent’s teammate (agent 2) immediately comprehends all agents’ identities; hence, agent 2’s confidence regarding agent 0 being a teammate rapidly approaches one. For the other two (agent 1 and agent 3), agent 2’s confidence levels rapidly approach zero. The confidence level curves during a round are illustrated in Fig. \ref{fig8} (between the red and purple vertical lines). 
    \item After both red 10 cards are played, the identities of all agents are exposed. Hence, all agents can identify their teammates with a confidence level that approaches one. Their confidence levels for their non-teammate approach zero naturally. As this is also illustrated in Fig. \ref{fig8} (right side of the purple vertical).
\end{enumerate}

\begin{figure*}[htbp]
\centering
    \subfigure[agent 0]{
    \begin{minipage}[t]{0.23\textwidth}
        \centering
        \includegraphics[width=1\textwidth]{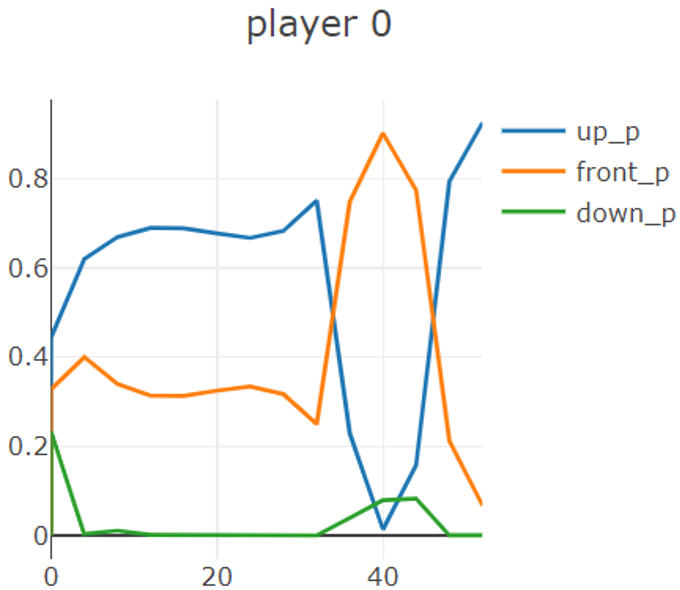}
    \end{minipage}}
    \subfigure[agent 1]{
    \begin{minipage}[t]{0.23\textwidth}
        \centering
        \includegraphics[width=1\textwidth]{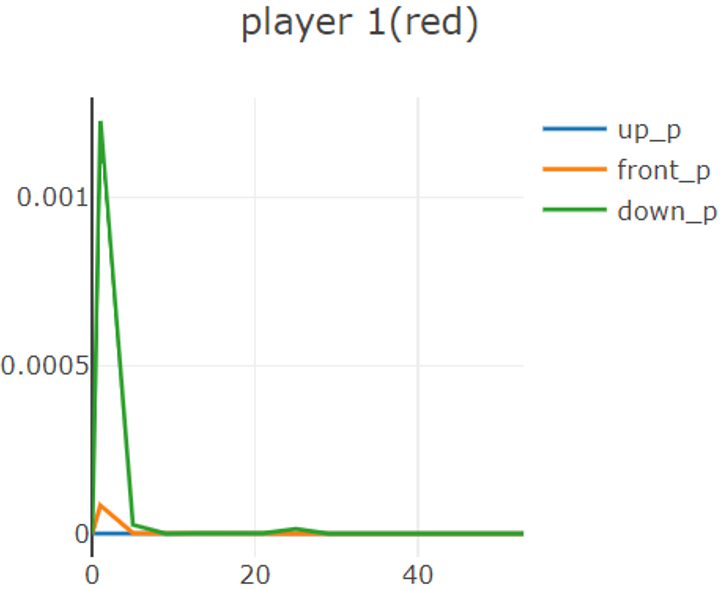}
        \end{minipage}}
    \subfigure[agent 2]{
    \begin{minipage}[t]{0.23\textwidth}
        \centering
        \includegraphics[width=1\textwidth]{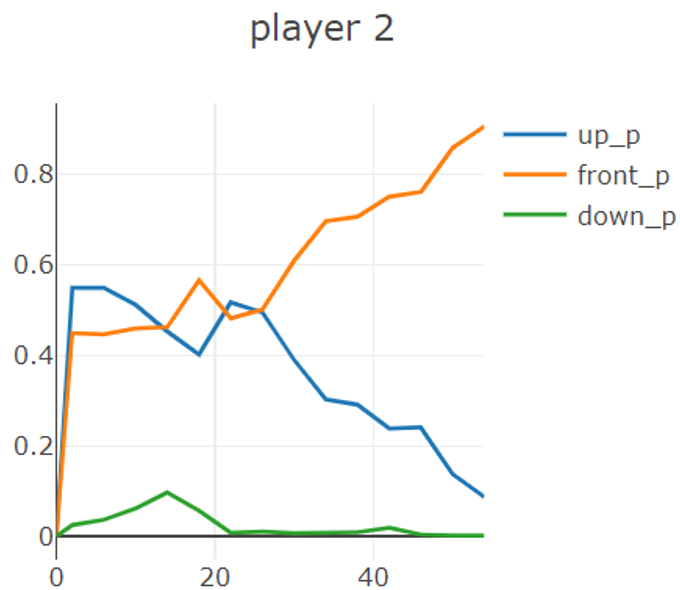}
    \end{minipage}}
    \subfigure[agent 3]{
    \begin{minipage}[t]{0.23\textwidth}
        \centering
        \includegraphics[width=1\textwidth]{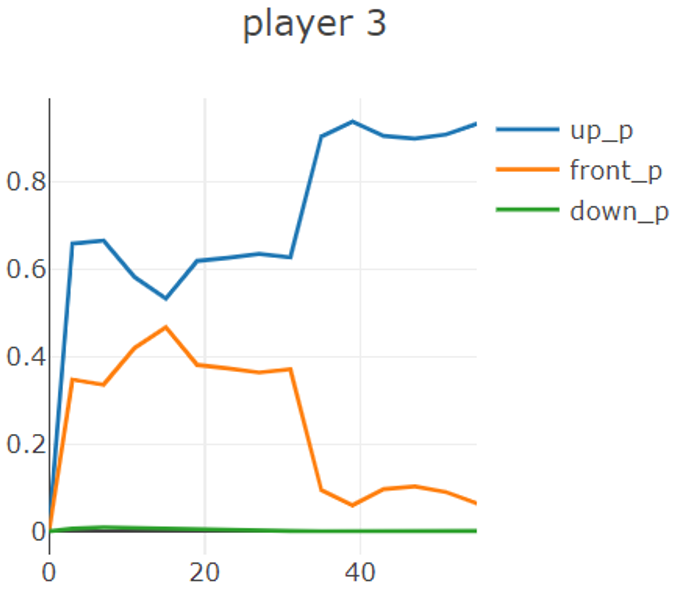}
    \end{minipage}}
    \caption{Confidence level curves during a round where a single agent holds two red 10 cards (agent 1 = landlord team).}
    \label{fig9}
\end{figure*}

\subsubsection{Partially Ambiguous Identities}
\begin{figure}[htbp]
\centerline{\includegraphics[width=0.5\textwidth]{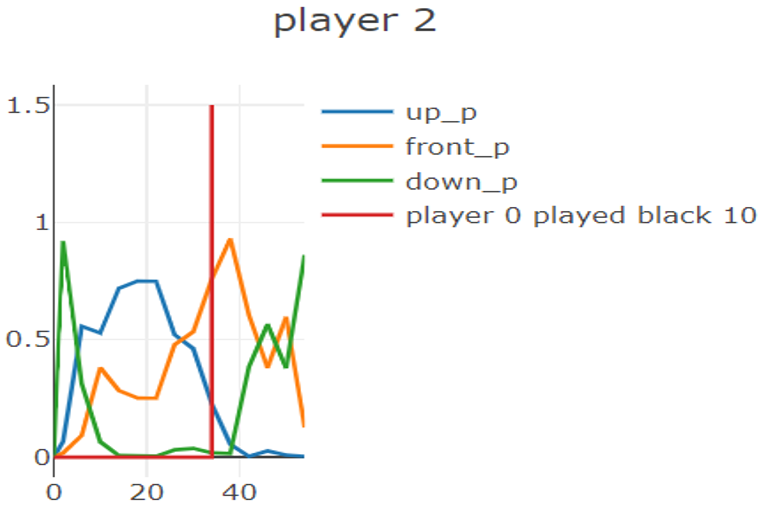}}
\caption{Confidence level curves of agent 2 (peasant). The red vertical represents the time step agent 0 plays the black card 10.}
\label{fig10}
\end{figure}

When the information about the red 10 card is not perfect, agents are unsure of the teams. Hence, the relation network must infer teammate identities by combining the information on historical actions and agents’ card-playing characteristics to observe other agents’ levels of cooperation and competition intention and make inferences. 
\begin{enumerate}
    \item As shown in Fig. \ref{fig8}, referring to scene 2 in the ``partially public identities’’ section, agent 0 still does not know the status of the other three agents. Therefore, the relation network must be invoked so that agent 0 can infer its teammate’s identity by compiling information on the other three agents’ cooperative and competitive behaviors.
    Over time, agent 0 will notice that agent 2 is behaving cooperatively, agent 1 and agent 3 are behaving competitively. Then, agent 0's confidence level of agent 2 increases, and the confidence level of agent 1 and agent 3 decreases with time gradually.
    \item Referring to scene 1 in the ``partially public identities’’ section, when two red 10 cards belong to a single agent and are not played (see Fig. \ref{fig9}), agents 0, 2, and 3 are unaware of their teammates’ identities. Hence, they must also invoke their relation network to ascertain that agent 1 is behaving competitively toward them, and decrease the confidence level of agent 1 gradually(agent 0’s down player, agent 2’s up player, agent 3’s front player).
    \item When agent 0 plays only one black 10 card, the confidence levels of the other landlord (peasant) agents who do (do not) hold a red 10 card to agent 0 noticeably decrease (increases). See Fig. \ref{fig10}. It demonstrates that the relation network can reasonably infer the agent’s identity by the information about the black card 10. 
\end{enumerate}

\subsubsection{Prejudice of the Relation Network}
With thousands of confidence-level curves, we noticed that when there is little information about the agents' identities, the active player seems to induce an increase in confidence in the up player, whereas a decrease in confidence is induced in the down player. This confidence pressure waveform follows the active player position as it rotates around the table: the down player plays cards just after the agent which directly suppresses the active agent, so the card-playing may be misunderstood as a confrontation; instead, the up player plays cards after two playing behind the agent, so the relation network observes the up player's competition intention insensitively. At the initial stage of Figs. \ref{fig8} and \ref{fig9} demonstrate this phenomenon, which unsurprisingly mimics the behaviors of real human players in similar situations.
As human owns a card-playing inclination that tends to suppress the down player and believe the up player.

\subsection{Analysis of the Danger Network}\label{sec5.3}
Here, we analyze the efficacy of the danger network during the gameplay described above.

\subsubsection{Partially Public Identities}
\begin{figure*}[htbp]
\centering
    \subfigure[agent 0]{
    \begin{minipage}[t]{0.23\textwidth}
        \centering
        \includegraphics[width=1\textwidth]{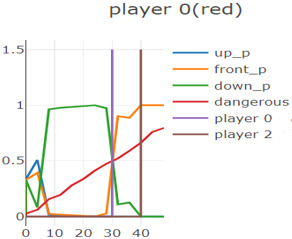}
    \end{minipage}}
    \subfigure[agent 1]{
    \begin{minipage}[t]{0.23\textwidth}
        \centering
        \includegraphics[width=1\textwidth]{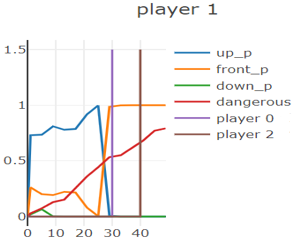}
        \end{minipage}}
    \subfigure[agent 2]{
    \begin{minipage}[t]{0.23\textwidth}
        \centering
        \includegraphics[width=1\textwidth]{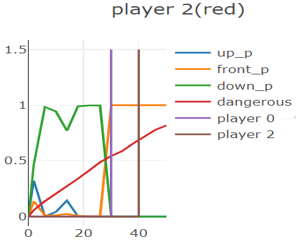}
    \end{minipage}}
    \subfigure[agent 3]{
    \begin{minipage}[t]{0.23\textwidth}
        \centering
        \includegraphics[width=1\textwidth]{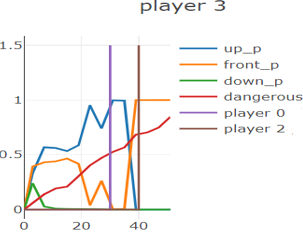}
    \end{minipage}}
    \caption{Confidence level and risk rate curves in a round where two agents (0 and 2) each hold one red 10 card.}
    \label{fig11}
\end{figure*}

When getting precise information about the red 10 card, the relation network accurately identifies some agents’ identities, and the confidence level tends to be polarized toward these agents. Thus, the confidence level of teammates (opponents) in these agents will be larger (less) than the risk rate evermore, and choose to cooperate (compete) with them. In this situation, the identification module accurately identifies the agents’ identities, and the danger network won’t hinder the accurate identification. As shown in Fig. \ref{fig11}, we use a red line to show the risk rate curve. After agent 0 plays a red 10 card (right side of the purple vertical), agent 2’s confidence level regarding agent 0 (front player) outweighs the following risk rates, and the confidence levels of agents 1 and 3 to agent 0 (agent 1's up player and agent 3's down player) are less than risk rate. After the two red 10 cards are played (right side of the brown vertical), the phenomenon is consistent with our explanation. 

\subsubsection{Partially Ambiguous Identities}
At the beginning of a game, everyone’s perceived risk rate is low. Each agent's confidence levels of other agents with ambiguous identities are larger than the risk rate. Hence, players tend to cooperate. Later in the game, however, the perceived risk rates grow to overshade the confidence levels until precise knowledge is gained by one or more agents. Hence, players tend to compete with agents with ambiguous identities to avoid potential opponents win the game.

\begin{figure*}[htbp]
\centering
    \subfigure[agent 0]{
    \begin{minipage}[t]{0.23\textwidth}
        \centering
        \includegraphics[width=1\textwidth]{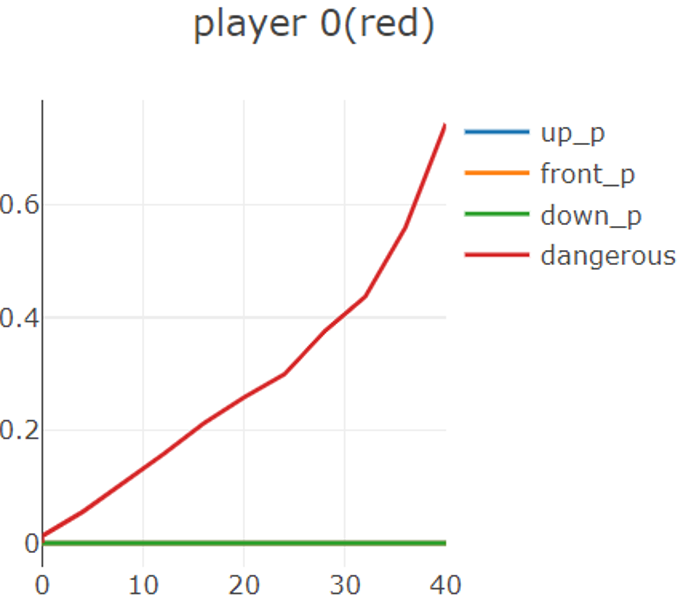}
    \end{minipage}}
    \subfigure[agent 1]{
    \begin{minipage}[t]{0.23\textwidth}
        \centering
        \includegraphics[width=1\textwidth]{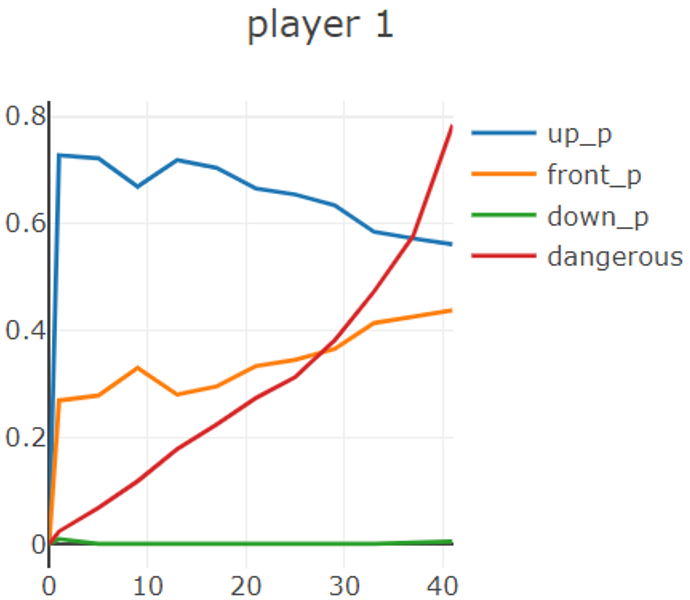}
        \end{minipage}}
    \subfigure[agent 2]{
    \begin{minipage}[t]{0.23\textwidth}
        \centering
        \includegraphics[width=1\textwidth]{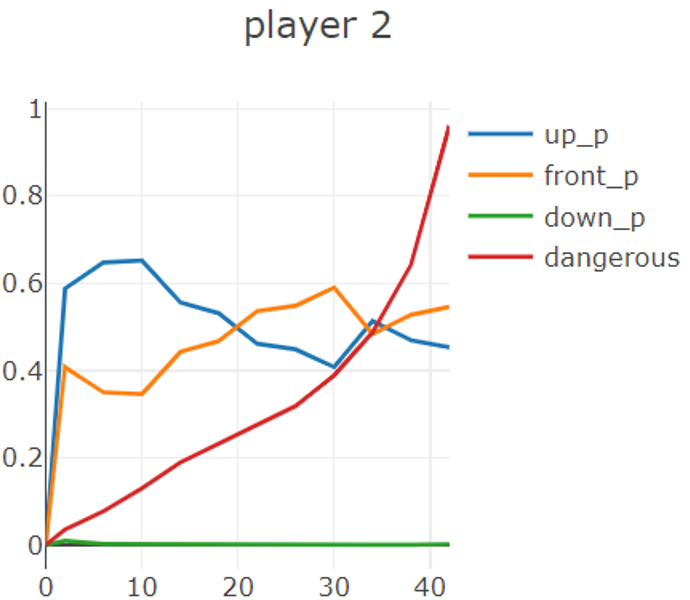}
    \end{minipage}}
    \subfigure[agent 3]{
    \begin{minipage}[t]{0.23\textwidth}
        \centering
        \includegraphics[width=1\textwidth]{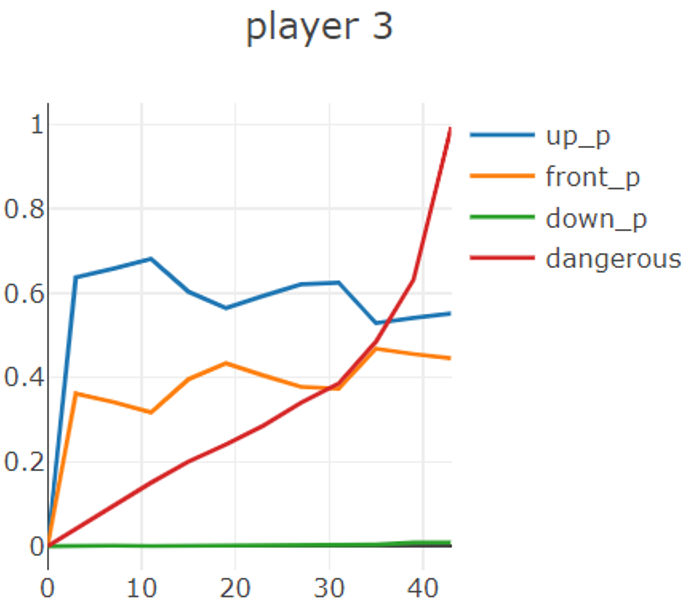}
    \end{minipage}}
    \caption{Confidence level and risk rate curves in a round where a single agent (0) holds both red 10 cards.}
    \label{fig12}
\end{figure*}

When the information about the red 10 card is not perfect and some agents’ identities are ambiguous, the danger network helps us mitigate the risk of inaccurate identification of the relation network. As in Fig. \ref{fig12}, agent 0 holds both red 10 cards and does not play them, so the other agents don’t know the others’ identities. At the beginning of the round, the confidence levels of agents 1, 2, and 3 to their up-player and front-player are larger than the risk rate. As time goes on, their risk rates grow and continue to overshadow their confidence levels. Hence, each of them competes with the other three agents to mitigate the risk of believing potential opponents.


\subsection{Ablation Study}
In this section, we present the ablation studies on the danger network and the identification module.
\subsubsection{Ablation Study on the Danger Network}
We conducted an ablation study to check the efficacy of the danger network, replacing it with a constant \begin{math}\nu\end{math}. The win rates of the complete IDRL are shown in Table \ref{table3}, note that the complete IDRL performs consistently better than the variant version without the danger network, regardless of large or small constant. Given a large constant risk rate, the identification module tends to assume other agents as opponents, which is much more vigilant during the early part of a round, and play became unreasonably suppressive of others. With a small constant risk rate, the identification module takes other agents as teammates and is unguarded near the end of a round, which may help the potential opponent win the game. Hence, the danger network was necessary for overall superior performance.

 In summary, the danger network dynamically calculates the risk rate, which mitigates the risk of inaccurate identification turn-by-turn. It further enables possible cooperation with diligent risk awareness.
\begin{table}[htbp]
\setlength{\abovecaptionskip}{0.05cm} 
\centering
\caption{IDRL w/ the danger network vs. IDRL w/o the danger network.} 
\begin{tabular}{l| r r r r}
	\hline
	   constant \begin{math}\nu\end{math}&0.2&0.4&0.6&0.8 \\ \hline
	  \textbf{\textit{Win Rate(\%)}}&53.6&54.7&54.6&53.4\\ \hline
\end{tabular}
\label{table3}
\end{table}

\subsubsection{Ablation Study on the Identification Module}\label{sec5.4}
We conduct an ablation study to assess the efficacy of the full identification module. After removing the identification module, we only use the policy module which is trained by a deep Monte-Carlo method. Then, we compare the performance as listed in Table \ref{table4}. The complete IDRL algorithm dominates the version without the identification module, which demonstrates the significance of the identification module. 

\begin{table}[htbp] 
\setlength{\abovecaptionskip}{0.05cm} 
\centering
\caption{IDRL vs. Monte-Carlo (the policy module).} 
\begin{tabular}{l| r }
	\hline
	   &\textbf{\textit{Win Rate(\%)}}\\ \hline
	   \textit{Monte-Carlo}&28.1 \\ \hline
	   \textit{IDRL}&71.9\\ \hline
\end{tabular}
\label{table4}
\end{table}

\section{Conclusion}\label{sec6}
To solve mixed cooperative and competitive games where the identities of agents are ambiguous and dynamic, we develop IDRL, a novel multi-agent reinforcement learning framework consisting of identification and policy modules that learns the policy with identification capability. The identification module consists of relation and danger networks, which it uses to identify agents’ identities and estimate the risk of making a mistake based on the task at hand and the reward function. The result was a safe and reliable cooperation--competition pattern of gameplay that ensures a balanced trade-off between maximum rewards and identification accuracy. In the chosen cooperation--competition pattern, we select the policy that can cooperate and compete with corresponding agents to do the decision-making. We applied an intrinsic reward method to continually update the identification module during gameplay.

In the future, we plan to explore additional research directions. For example, a potential direction could be identifying more complex relationships (instead of only two types) between agents in a complex environment. It might be related to causal relationship discovery.
Beyond that, in real-world problems, there are some agents with superior knowledge, who would like to hide their identities. Therefore, how the relation network should work in more complicated circumstances needs to be further investigated. 
\\

\backmatter
\noindent{\textbf{Acknowledgments.}} We acknowledge funding in support of this work from the Project supported by the Key Program of the National Natural Science Foundation of China (Grant No.51935005), Basic Research Project (Grant No.JCKY20200603C010), and supported by Natural Science Foundation of Heilongjiang Province of China (Grant No.LH2021F023), as well as supported by Science and Technology Planning Project of Heilongjiang Province of China (Grant No.GA21C031).

\begin{appendices}

\section*{Appendix A: Red-10 Game Rules.}
{\bf Deck:} Red-10 game is played with a standard 52-card deck comprising 13 ranks in each of the four suits: clubs, diamonds, hearts, and spades. Each suit series is ranked from top to bottom as 2,A,K,Q,J,10,9,8,7,6,5,4,3. 

{\bf Cards combination categories:} Similar to the Doudizhu, there are rich card combination categories in Red-10 as follows.
\begin{itemize}
    \item Solo: Any individual card, ranked according to its face rank.
    \item Pair: Any pair of identically ranked cards, ranked according to its face rank.  
    \item Trio: Any three identically ranked cards, ranked according to its face rank.
    \item Trio with solo: Any three identically ranked cards with a solo, ranked according to the trio.
	\item Trio with pair: Any three identically ranked cards with a pair, ranked according to the trio.
    \item Solo chain: No fewer than five consecutive card ranks, ranked by the lowest rank in the chain.  
    \item Pairs chain: No fewer than three consecutive pairs, ranked by the lowest rank in the chain.
    \item Airplane: No fewer than two consecutive trios, ranked by the lowest rank in the combination.
    \item Airplane with small wings: No fewer than two consecutive trios, with additional cards having the same amount of trios, ranked by the lowest rank in the chain of trios.
    \item Airplane with large wings: No fewer than two consecutive trios with additional pairs having the same amount of trios, ranked by the lowest rank in the chain of trios. 
    \item Four with two single cards: Four cards with equal rank with two individual cards, ranked according to the four cards.
    \item Four with two pairs: four cards with equal rank, with two pairs, ranked by the four cards.
    \item Bomb: Four cards of equal rank. 
\end{itemize}

Red-10 includes two phases as follows.
\begin{enumerate}
    \item Dealing: A shuffled deck of 52 cards is randomly dealt to four players in turn, equally.
    \item Card-playing: For players play cards in turn; the first plays any category. The next player must play cards of the same category with a higher rank or bomb; otherwise, they can pass on their turn. If three consecutive agents pass, the fourth player can play any category. The game ends when any player runs out of cards.
\end{enumerate}

{\bf Winner:} Players holding a red 10 card are on the ``Landlord team,’’ and the others on the ``Peasant.’’ The first team with a player who runs out of cards wins.
\section*{Appendix B: Detailed Input Data}
In Red-10 game environment, the detailed input data of the Q action-value function, the relation network, and the danger network are listed as follows tables (Table \ref{table5} and Table \ref{table6}).
\begin{table}[htbp]
\centering
\caption{Input Data of the Q Action-Value Function.} 
\begin{tabular}{l| l r}
	\hline
	   &Feature &Size \\ \hline\hline
	   action& a legal action &52\\ \hline
	   state& hand cards&52\\
	   &the union of the other three agents' hand cards&52\\
	   &the cards played most recently&52\\ 
	   &the cards up player played most recently&52\\
	   &the cards front player played most recently&52\\
	   &the cards down player played most recently&52\\
	   &the union of the up player's historical actions&52\\
	   &the union of the front player's historical actions&52\\
	   &the union of the down player's historical actions&52\\
	   &the number of up player's hand cards&13\\
	   &the number of front player's hand cards&13\\
	   &the number of down player's hand cards&13\\
	   &the most recent 20 cards played&\begin{math}
	   5\times208
	   \end{math}\\ 
\end{tabular}
\label{table5}
\end{table}

\begin{table}[htbp]
\centering
\caption{Input Data of the Relation and Danger networks.} 
\begin{tabular}{l| l r}
	\hline
	   &Feature &Size \\ \hline\hline
	   state&hand cards&52\\
	   &the union of the other three agents' hand cards&52\\
	   &the cards up player played most recently&52\\
	   &the cards front player played most recently&52\\
	   &the cards down player played most recently&52\\
	   &the union of the up player's historical actions&52\\
	   &the union of the front player's historical actions&52\\
	   &the union of the down player's historical actions&52\\
	   &the number of up player's hand cards&13\\
	   &the number of front player's hand cards&13\\
	   &the number of down player's hand cards&13\\
	   &the suits of card 10 up player's played&4\\
	   &the suits of card 10 front player's played&4\\
	   &the suits of card 10 down player's played&4\\
	   &the suits of card 10 ever in hand&4\\
	   &the union of card 10 suit in other agents' hands&4\\
	   &the most recent 20 cards played&\begin{math}
	   5\times208
	   \end{math}\\ 
\end{tabular}
\label{table6}
\end{table}




\end{appendices}


\bibliography{sn-article}


\end{document}